\documentclass[runningheads]{llncs}
\usepackage[T1]{fontenc}
\usepackage{cite} 
\usepackage{graphicx,verbatim}
\usepackage{multirow}
\usepackage{multicol}
\usepackage{amsmath}
\usepackage{amsfonts}
\usepackage{fontawesome}

\usepackage{svg}

\begin{document}
\title{
 Temporal Representation Learning of Phenotype Trajectories for pCR Prediction in Breast Cancer
}

\author{Ivana Janíčková\inst{1,2}\and
Yen Y. Tan \inst{3}\and
Thomas H. Helbich\inst{1}\and
Konstantin Miloserdov\inst{1,2,5}\and
Zsuzsanna Bago-Horvath\inst{4}\and
Ulrike Heber\inst{4}\and
Georg Langs\inst{1,2,5}
}
\authorrunning{I. Janíčková et al.}
%
\institute{Computational Imaging Research Lab, Department of Biomedical Imaging and Image-guided Therapy, Medical University of Vienna, Austria 
\and 
Comprehensive Center for Artificial Intelligence in Medicine, Medical University of Vienna, Austria \and
Department of Obstetrics and Gynecology, Medical University of Vienna, Austria \and
Department of Pathology, Medical University of Vienna, Austria \and
Christian Doppler Laboratory for Machine Learning Driven Precision Imaging, Department of Biomedical Imaging and Image-guided Therapy, Medical University of Vienna, Austria\\
\email{ivana.janickova@meduniwien.ac.at}, 
\email{georg.langs@meduniwien.ac.at} 
\texttt{\url{https://www.cir.meduniwien.ac.at}}
}

\titlerunning{Temporal Representation Learning for predicting pCR}

\maketitle              
\begin{abstract}
Effective therapy decisions require models that predict the individual response to treatment. This is challenging since the progression of disease and response to treatment vary substantially across patients. Here, we propose to learn a representation of the early dynamics of treatment response from imaging data to predict pathological complete response (pCR) in breast cancer patients undergoing neoadjuvant chemotherapy (NACT). The longitudinal change in magnetic resonance imaging (MRI) data of the breast forms trajectories in the latent space, serving as basis for prediction of successful response. The multi-task model represents appearance, fosters temporal continuity and accounts for the comparably high heterogeneity in the non-responder cohort.In experiments on the publicly available ISPY-2 dataset, a linear classifier in the latent trajectory space achieves a balanced accuracy of 0.761 using only pre-treatment data ($T_0$), 0.811 using early response ($T_0+T_1$), and 0.861 using four imaging time points ($T_0 \rightarrow T_3$). The full code can be found here: \url{https://github.com/cirmuw/temporal-representation-learning}
\keywords{Temporal representation learning, Self-supervised learning, Breast Cancer}
\end{abstract}

\section{Introduction}
Pathological complete response (pCR) to neoadjuvant chemotherapy (NACT) of breast cancer is a key marker of success determined on the basis of tissue resected during surgery~\cite{li2020predicting}. Predicting pCR by assessing early response dynamics can steer treatment decisions. Even after concluded NACT, it may inform important choices such as forgoing surgery in case of expected pCR, given sufficient prediction reliability. While single time point observations lack information on subtle, dynamic changes to treatment~\cite{jing2024prediction,qu2020prediction}, longitudinal imaging may capture changes associated with individual treatment efficacy or disease progression. 

Here, we propose a multi-task model to learn trajectory representations of imaging features observed during treatment. We show how a simple classifier can use this representation to predict future pCR with high accuracy. Our approach addresses the challenge of high inter-label similarity~\cite{ konz2024effect} and relatively substantial inter-individual variability not associated with treatment response. The dynamics of early response enables better prediction, while multi-task representation learning accounts for the response heterogeneity.

\paragraph{Related work}\label{sec:relatedwork}

Prior efforts to predict pCR in breast cancer imaging have used radiomics-based \cite{li2020predicting,huang2023longitudinal,li202018} and deep learning-based approaches \cite{zhang2024m2fusion,jing2024prediction,joo2021multimodal,qu2020prediction,bulut2023prediction,comes2021early,dammu2023deep}. These methods largely focus on single \cite{joo2021multimodal,li202018,bulut2023prediction,dammu2023deep} or two time point predictions \cite{zhang2024m2fusion,huang2023longitudinal,qu2020prediction}, limiting their ability to capture the full temporal dynamics of tumor progression.  Although some studies incorporate multiple time points for pCR prediction \cite{jing2024prediction,comes2021early}, only  \cite{jing2024prediction} explicitly models temporal relationships using an LSTM layer \cite{hochreiter1997long}. Using three imaging time points, the model achieved AUC = 0.706, Sensitivity = 0.483 and Specificity = 0.773 \cite{jing2024prediction}. Learning temporal relationships by creating individualized treatment progression trajectories or leveraging the learned temporal relationships to improve single time point predictions remains unexplored. The publicly available ISPY-2 dataset consists of series of MRIs taken before and during NACT~\cite{Li2022_ISPY2,newitt2021acrini}. It provides an opportunity to address these limitations by modelling temporal dynamics more effectively. 
Existing methods typically classify patients as responders (positive) or non-responders (negative) \cite{thrasher2024te,zhang2024m2fusion,yue2022mldrl}. For instance, \cite{zhang2024m2fusion} pre-trains a pCR prediction model by clustering pre- and post-NACT non-pCR images (assuming minimal change during NACT) while separating pre- and post-NACT pCR images (assuming greater change). However, this binary framework overlooks the heterogeneity among non-responders, including partial responders. The classification performance of this approach resulted in a test set AUC of 0.695 for the binary pCR label \cite{zhang2024m2fusion}.

\paragraph{Contribution}

We propose a representation learning approach for image trajectories observed during treatment. We assume that, over time, responders change similarly, while non-responders are more heterogeneous since they include both partial responders and non-responders. The multi-task model learns to embed appearance change, while fostering temporal continuity by a dynamic-margin triplet loss adapted to nuanced temporal relationships. It aligns trajectories of responders to identify common temporal dynamics associated with successful treatment response, while accounting for the heterogeneity within the non-responder group. It avoids label-driven loss functions for non-responders and instead identifies response-specific patterns by aligning positive-outcome trajectories. A multi-task attention mechanism (MTAN) \cite{liu2019end} enables the focus on feature changes associated with disease and treatment as opposed to comparably static inter individual variability of anatomy.

Experiments on the multi-center ISPY-2 dataset~\cite{Li2022_ISPY2,newitt2021acrini} demonstrate that the learned representation enables high prediction accuracy with a linear classifier, even with a single or only few time points. The MRI-based classification performance surpasses the state of the art as reported in \cite{zhang2024m2fusion,jing2024prediction}.

To our knowledge, this is the first pre-training method designed to (1) explicitly distinguish responder vs. non-responder progression patterns and (2) encode temporal relationships within ISPY-2. Unlike \cite{jing2024prediction} our approach directly learns temporal dynamics for improved treatment progression modeling. Additionally, it leverages learned trajectories to refine single time point predictions, such as early pCR prediction at $T_0$, enhancing its ability to capture treatment response patterns.

\begin{figure}[t]
    \centering
    \includegraphics[width=280pt]{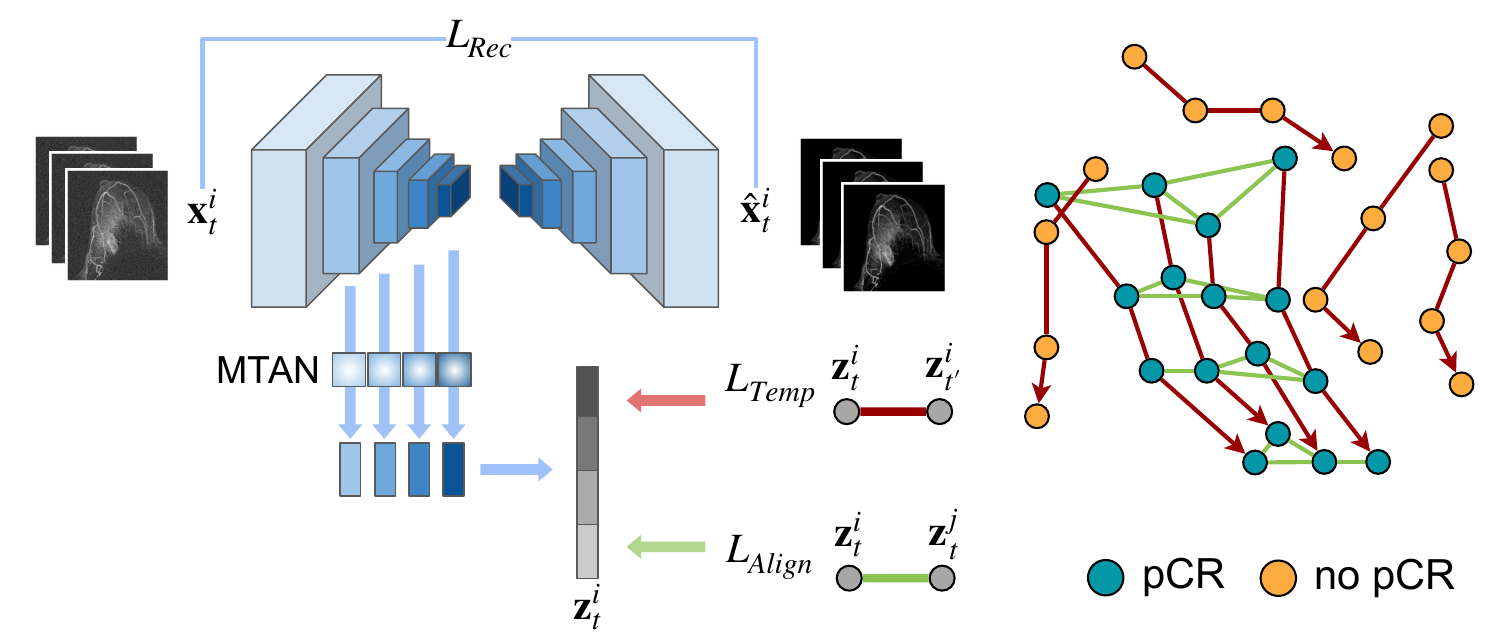}
    \caption{\textbf{Method Overview} Multi-task representation learning balances reconstruction performance  $L_{Rec}$ with temporal continuity of trajectories $L_{Temp}$, and alignment of changes in responders $L_{Align}$. A U-shaped denoising network extracts multi-scale features via its encoder. An MTAN module \cite{liu2019end} steers attention across these tasks. The resulting trajectory representations are used for pCR prediction with a linear classifier.
    }
    \label{fig:abstract}
\end{figure}

\section{Method} 
 
For each patient $i$, we observe a time-series of images $\mathbf{x}^i_t$ for $t=0,\dots, T$ acquired before ($t=0$), during, and after treatment ($t=T$), and a label $y^i\in\{0,1\}$, where \(1\) denotes a positive outcome (achieving pCR) and \(0\) a negative outcome (not achieving pCR). The multi-task training model for representing images integrates three loss functions: a triplet loss \cite{schroff2015facenet} with dynamic margin for temporal modeling~$L_{Temp}$, a cosine similarity loss for responder alignment~$L_{Align}$, and a reconstruction loss for robust image feature learning~$L_{Rec}$. 

A set of two distinct augmented examples is generated for each training example:  
\( A_1(\mathbf{x}^i_t) \) and \( A_2(\mathbf{x}^i_t) \).  
The final multiscale representations, derived from the encoder \( f \), are then given by:  
\(\mathbf{z}^i_t = f(A_1(\mathbf{x}^i_t))\), \(\mathbf{\bar{z}}^i_t = f(A_2(\mathbf{x}^i_t))\) as shown in Fig. \ref{fig:abstract}.
\\Our model is a U-shaped encoder-decoder network, where the multi-scale representations are generated from the encoder. The representation tensors are normalized to a unit hypersphere and were used to predict the pCR in patients.

\subsection{Constructing Multi-scale Visual Representations}
We train a U-shaped encoder-decoder network to obtain a multi-scale representation of images, while at the same time fostering temporal continuity and alignment of trajectories observed in responders (Fig. \ref{fig:abstract}). Fine-grained features are extracted from the encoder’s multi-scale feature maps. These are then pooled, projected and concatenated into a multi-scale representation tensor $\mathbf{z}$. To ensure that the extracted features are anatomically relevant, we incorporate a reconstruction task:
\begin{equation}\label{eq:6}
    L_{Rec} = \sum_{i \in N}\sum_{t \in T}E(\mathbf{x}^i_t, \mathbf{\hat{x}}^i_t)
\end{equation}
Here, \(\mathbf{x}^i_t\) is the target input example and \(\mathbf{\hat{x}}^i_t\) is the denoised reconstruction generated from the noise-augmented input \(A_1(\mathbf{x}^i_t)\). The loss function $E$ quantifies the reconstruction error using the mean squared error. 

\subsection{Learning Temporal Relationships}
To adapt the triplet loss \cite{schroff2015facenet} for learning patient-level temporal relationships, we define an anchor-positive pair as representations of two different views at the same time point $t$: \(a = \mathbf{z}^{i}_t\), \(p = \mathbf{\bar{z}}^{i}_t\). The negative point is the image representation from the same patient at a different time point  \(n = \mathbf{z}^{i}_{t'}\) where \(t' \neq t\). Instead of a fixed margin, we use a dynamically changing margin \(m\), based on the relative difference between \(t\) and \(t'\). Additionally, we replace the standard distance metric with negative cosine similarity \(d\). The final triplet loss is defined as follows, with \(N\) denoting the total number of instances in a batch:

\begin{equation}
    L_{Temp} = \sum_{i \in N}\sum_{t \in T}\sum_{t' \in T} max(d(\mathbf{z}^{i}_t , \mathbf{\bar{z}}^{i}_t) - d(\mathbf{z}^{i}_t, \mathbf{z}^{i}_{t'}) + m, 0)
\end{equation}

\subsection{Learning Shared Patterns of Change in Responders}

In order to learn the shared patterns in responders' temporal trajectories, their representations are aligned by establishing correspondences in the latent space. The alignment is defined for two patients (\(i, j\)) with \(y^i = 1\) and \(y^j = 1\) and their corresponding representation tensors (\(\mathbf{z}^i_t,\mathbf{z}^j_t\)). The objective is to learn population-level response patterns by minimizing the distance between \(\mathbf{z}^i_t\) and \(\mathbf{z}^j_t\). The alignment loss is then defined for pairs representation as:
\begin{equation}\label{eq:5}
    L_{Align} = \sum_{i \in N} \sum_{j \in N}\sum_{t \in T} \mathbb{I}(y_i = 1 \land y_j = 1) d(\mathbf{z}^{i}_t, \mathbf{z}^{j}_t))
\end{equation}

Here, \(\mathbb{I}\) is an indicator function that ensures that the loss is only computed for pairs of positive examples. 
 The expression $d(\mathbf{z}^{i}_t, \mathbf{z}^{j}_t)$ represents the negative cosine similarity between the two representations.

\subsection{Overall Loss Function}
The objective of the pre-training phase is to optimize the combined loss function:
\begin{equation}\label{eq:total}
   L_{ART} =
    \begin{cases} 
        L_{Align} + L_{Rec} + L_{Temp}, & \text{if } y = 1, \\
        L_{Rec} + L_{Temp}, & \text{otherwise}.
    \end{cases}
\end{equation}
The combined loss function ensures that the positioning of negative-outcome examples ($y = 0$) in the latent space is influenced only by the reconstruction and temporal components. In contrast, positive-outcome examples are further aligned at a population level through the supervision (\(L_{Align}\), Fig. \ref{fig:abstract}).

\subsection{Feature Masking in Multi-Task Learning}
We incorporate a learnable attention mask inspired by the MTAN module  \cite{liu2019end} to emphasize temporal changes and response-specific patterns in feature maps. It balances shared feature learning ($L_{Rec}$) with task-specific details ($L_{Align}+L_{Temp}$), refining representations to capture spatio-temporal changes during treatment.

\section{Experiments and Results}

\subsubsection{ISPY-2 Dataset} 
 We used 585 patients from the public ISPY-2 dataset \cite{Li2022_ISPY2, newitt2021acrini} with complete MRI scans at four NACT time points and all three DCE-derived maps: early enhancement (\(PE_{\text{early}}\), 120--150 sec post-contrast), late enhancement (\(PE_{\text{late}}\), $\sim$450 sec), and signal enhancement ratio (\(SER = \frac{PE_{\text{early}}}{PE_{\text{late}}}\)), enabling consistent longitudinal comparisons. These features capture contrast washout dynamics, offering insights into tumor biology and vascular properties \cite{el2009dynamic}. Within this cohort, the proportion of patients achieving pCR is 33 \%. To reduce memory usage, we generated axial-plane maximum intensity projections (MIPs) of the three DCE-derived volumes. The dataset was split into 70\% training-, 10\% validation-, and 20\% test sets, stratified by pCR label. All volumes were resized to 256×256×256 and intensity-normalized to [0,1] before MIP generation.

\subsubsection{Implementation Details}  
The model is built on a UNet backbone \cite{falk2019u} using MONAI’s BasicUNet \cite{cardoso2022monai}, initialized with features argument set to [16, 32, 64, 128, 256, 32]. The encoder-decoder structure was used for reconstruction (Fig. \ref{fig:abstract}), while multi-scale encoder features were concatenated for temporal and response learning. We incorporated MTAN’s masking strategy \cite{liu2019end} to selectively refine relevant feature maps, ensuring better alignment with temporal and response dynamics. A two-layer MLP projector was then applied, resulting in a final feature size of 480.
Pre-training was conducted using Adam optimizer \cite{kingma2014adam} with a learning rate of 0.0001 and a batch size of 32 for 100 epochs. The triplet loss margin was dynamically set by encoding MRI time points in range of \([0,1]\) with a step size of 0.25. For comparison, we used \(\mathcal{L}_{TESSL}\), a time- and event-aware SSL strategy \cite{thrasher2024te} introduced at MICCAI 2024, which, like our approach, incorporates both temporal and supervised signals during pre-training. Gradient accumulation was applied over 8 iterations with a batch size of 16 to simulate an effective batch size of 128, with Adam (learning rate = 0.15) as the optimizer. All models were pre-trained on full time-series data ($T_0 \rightarrow T_3$).  

\subsubsection{Evaluation measures}
For evaluation, we applied linear classifier to the frozen pre-trained features using \texttt{sklearn}’s LogisticRegression. Performance was assessed on the baseline time point ($T_0$) and the full time-series ($T_0 \rightarrow T_3$) over 10 runs, reporting the mean and standard deviation for the area under the receiver operating characteristic curve (AUROC), the area under the precision-recall curve (PRAUC) \cite{davis2006relationship}, and balanced accuracy. We further analyzed early response ($T_0 + T_1$) using sensitivity, specificity, positive predictive value (PPV), and negative predictive value (NPV). Model and parameter selection were performed on the validation set, results are reported for the test set.

\subsection{Results} 

\begin{figure}[t]
    \centering
    \includegraphics[width=\linewidth]{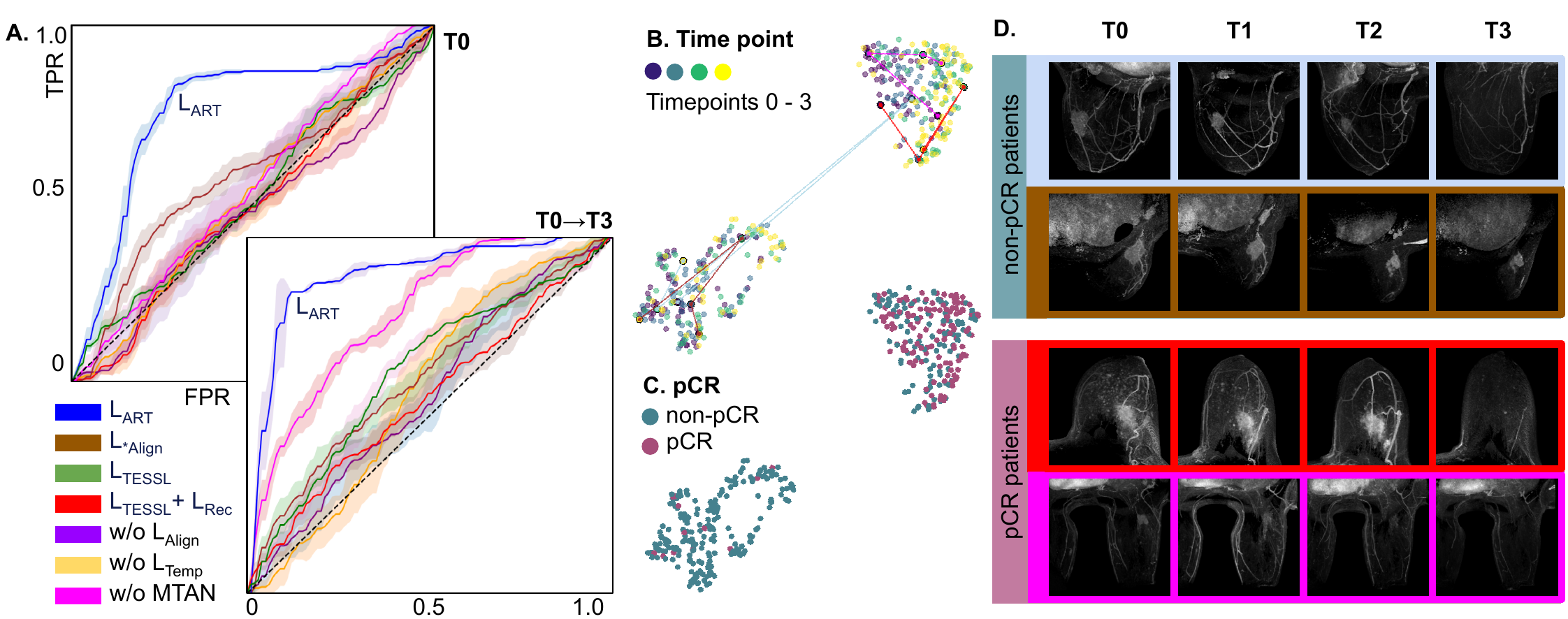}
    \caption{\textbf{(A)} ROC for pCR predictions of different models. \textbf{(B)} UMAP projection of the test set data, colored by time point label, with plotted trajectories for two non-pCR patients and two pCR patients. \textbf{(C)} Same as B, colored by pCR label. \textbf{(D)} Image time-series for four patients, with visualized trajectories across all time points. Frame colors correspond to the trajectory colors.}
    \label{fig:results}
\end{figure}

\begin{table}[b]
    \centering
    \caption{Linear evaluation results for the pCR prediction for early time point ($T_0$) and entire time-series ($T_0 \rightarrow T_3$) comparing a baseline approach $L_{TESSL}$ and $L_{TESSL}$ and $L_{Rec}$, ablated versions of the proposed approach (exluded components marked with w/o), and the proposed approach $L_{ART}$.}
    \label{tab:results}
    {\fontsize{8}{10}\selectfont
   \begin{tabular}{|l|@{\hspace{4pt}}c@{\hspace{4pt}}c@{\hspace{4pt}}c@{\hspace{4pt}}c@{\hspace{4pt}}c@{\hspace{4pt}}c|} 
        \hline
        \multirow{2}{*}{Method} & \multicolumn{2}{c}{AUROC} & \multicolumn{2}{c}{PRAUC} & \multicolumn{2}{c|}{Balanced Acc} \\ 
                                 & $T_0$ & $T_0\rightarrow T_3$ & $T_0$ & $T_0\rightarrow T_3$ & $T_0$ & $T_0\rightarrow T_3$ \\ 
        \hline
        \(L_{TESSL}\)              &0.625\(\pm .01\) &0.556\(\pm.01\) & 0.367\(\pm.01\)  &0.449\(\pm.01\)  & 0.526\(\pm.01\) &0.565\(\pm.01\) \\
        \(L_{TESSL} + L_{Rec}\)    &0.507\(\pm.02\)  &0.556\(\pm.01\) & 0.321\(\pm.01\)  &0.413\(\pm.03\)  & 0.472\(\pm.02\) & 0.556\(\pm.03\) \\
        \hline
        \textrm{w/o MTAN}        &0.558\(\pm.02\) &0.783\(\pm.01\) &0.359\(\pm.01\)  &0.613\(\pm.01\) &0.528\(\pm.03\) &0.697\(\pm.02\) \\
        \textrm{w/o} \(L_{Temp}\)              &0.495\(\pm .02\) &0.562\(\pm.03\) & 0.322\(\pm.02\)  &0.346\(\pm.02\)  &0.492\(\pm.03\) &0.518\(\pm.04\) \\
        \textrm{w/o} \(L_{Align}\)              &0.575\(\pm.02\)  &0.603\(\pm.01\) &0.388\(\pm.02\)  &0.442\(\pm.02\) & 0.593\(\pm.02\) &0.567\(\pm.02\) \\
        + \(L_{Align}^*\)         &0.475\(\pm.02\) &0.554\(\pm.03\) &0.312\(\pm.01\)  &0.356\(\pm.02\)  & 0.514\(\pm.02\) &0.528\(\pm.03\) \\

        \hline
        \(L_{ART}\)                & \textbf{0.764\(\pm\textbf{.01}\)} &\textbf{0.892\(\pm\textbf{.01}\)} & \textbf{0.565}\(\pm\textbf{.02}\)  & \textbf{0.746\(\pm\textbf{.03}\)} & \textbf{0.761\(\pm\textbf{.01}\)} &\textbf{0.861\(\pm\textbf{.01}\)} \\
        \hline
    \end{tabular}}
\end{table}

\subsubsection{Comparison with baseline method}
We compared our approach with the state-of-the-art model~\cite{thrasher2024te} and two pre-training strategies: baseline $L_{TESSL}$, $L_{TESSL}+L_{Rec}$, and our proposed $L_{ART}$ (Eq. \ref{eq:total}). Our method consistently outperformed the baseline across all metrics (Table \ref{tab:results}, Fig. \ref{fig:results}.A), achieving AUCROC of 0.892, PRAUC of 0.746, and Balanced Accuracy of 0.861 with the full time-series, and AUCROC of 0.764, PRAUC of 0.565, and Balanced Accuracy of 0.761 using only $T_0$ images. Bonferroni-corrected paired t-tests showed statistically significant differences (p<0.001) for all metrics and evaluation time points ($T_0$, $T_0 \rightarrow T_3$).

\begin{figure}[t]
    \centering
    \includegraphics[scale=0.30]{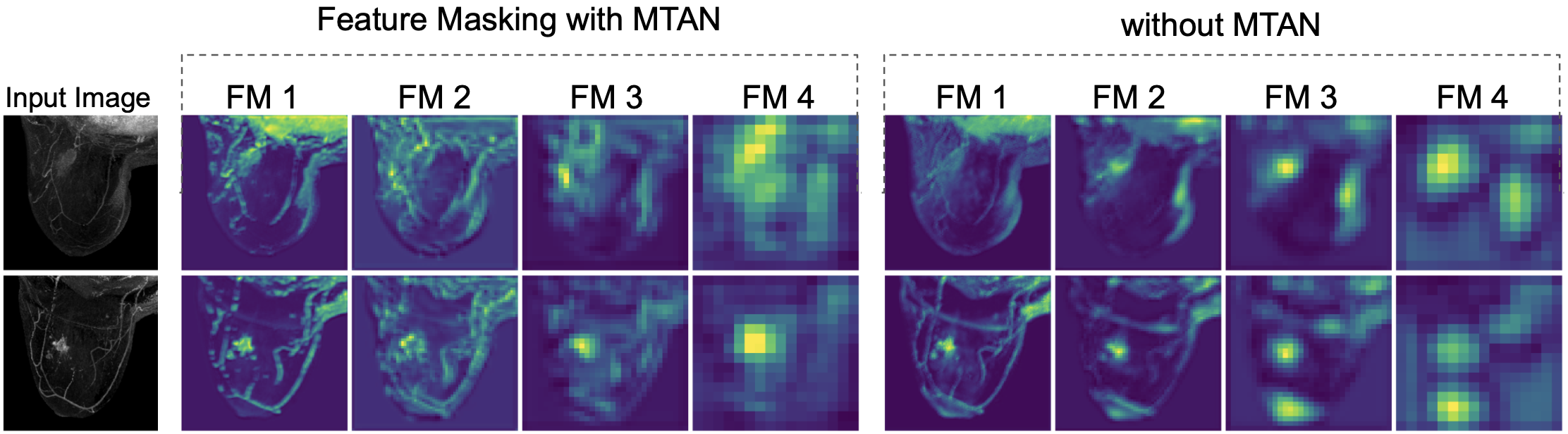}
    \caption{Comparison of feature maps (FM) extracted from encoder trained with and without the MTAN module. Four feature maps visualize different levels of the encoder.}
    \label{fig:mtan}
\end{figure}

\subsubsection{Ablation Study}  
We conducted an ablation study to assess the contributions of individual loss terms in our combined loss function \(L_{\text{ART}}\) (Table \ref{tab:results}, Fig. \ref{fig:results}.A). Three key variations were examined: (1) removing the temporal loss (w/o \(L_{\text{Temp}}\)), (2) removing the alignment loss (w/o \(L_{\text{Align}}\)), and (3) applying the alignment loss across both pCR and non-pCR patients instead of exclusively to pCR cases $(L^*_{Align})$. Across all measures, all ablations resulted in a performance decrease. In addition, we evaluated the impact of the MTAN module \cite{liu2019end}. As shown in Table \ref{tab:results}, removing MTAN significantly reduced performance, confirming its role in enhancing feature informativeness for pCR classification. Fig.\,\ref{fig:mtan} illustrates how MTAN masking improves the attention focus on tumor regions. Lastly, we visualised the latent space of our model, highlighting temporal (Fig. \ref{tab:results}.B) and outcome labels (Fig. \ref{tab:results}.C).

\subsubsection{Early Response Prediction}
Further analysis was performed to evaluate the prediction task for early response when only part of the time-series is available ($T_0+T_1$) as shown in Table \ref{tab:earlypred}. Adding $T_1$ improved performance across all metrics compared to pre-NACT prediction. Specificity and PPV reached values of 0.853 and 0.721, respectively, nearly matching those of the full time-series ($T_0\rightarrow T_3$). This demonstrates the feasibility of predicting treatment outcomes from early response dynamics and the benefit of limited temporal information compared to static pre-treatment data ($T_0$).

\begin{table}[t]
    \centering
    \caption{Comparison of prediction accuracy for data at pre-treatment ($T_0$), early response $T_0+T_1$, and full treatment timeline before surgery $T_0\rightarrow T_3$.}
    \label{tab:earlypred}
    {\fontsize{8}{10}\selectfont
    \begin{tabular}{|l|@{\hspace{2pt}}c@{\hspace{2pt}}c@{\hspace{2pt}}@{\hspace{1.5pt}}c@{\hspace{2pt}}c@{\hspace{2pt}}@{\hspace{2pt}}c@{\hspace{2pt}}c@{\hspace{2pt}}c@{\hspace{2pt}}|}
        \hline
        T &AUROC   &PRAUC  &Bal. Acc &Sensitivity &Specificity &PPV &NPV \\
                                  
        \hline
        $T_0$        &0.764$\pm .01$ &0.565$\pm .02$ &0.761$\pm .01$ & 0.731$\pm .02$ &0.762$\pm .02$  &0.603$\pm .02$ &0.852$\pm .01$\\
        $T_0+T_1$     & 0.802$\pm .01$ &0.649$\pm .02$ &0.811$\pm .02$ &0.769$\pm .04$ &0.853$\pm .01$  &0.721$\pm .02$ &0.883$\pm .02$ \\
        $T_0\rightarrow T_3$     &\textbf{0.892$ \pm\textbf{.01}$} &\textbf{0.746$ \pm\textbf{.03}$} &\textbf{0.861$\pm \textbf{.01}$} &\textbf{0.846$\pm \textbf{.00}$} &\textbf{0.876$\pm\textbf{.02}$} &\textbf{0.772$\pm\textbf{.02}$} &\textbf{0.920$\pm\textbf{.01}$} \\ 
        \hline
    \end{tabular}}
\end{table}

\section{Discussion}
We propose a novel method that captures the temporal phenotypic dynamics of treatment response. It learns to represent response-specific patterns in serial MRI of BC patients undergoing NACT. The multi-task model generates individual temporal trajectories, aligning behavior in responders and representing image appearance using a joint loss function $L_{ART}$ balanced with an MTAN attention masking mechanism.

Comparative results underscore the contribution of the individual components of $L_{ART}$. Removing the temporal term $L_{Temp}$ leads to performance degradation, likely due to representational collapse. Omitting the responder alignment term $L_{Align}$ results in poor linear probing performance due to a lack of supervision during pre-training. At the same time, accounting for the heterogeneity of the non-responder group is crucial, as demonstrated by the drop in performance when aligning trajectories within both the responder- and non-responder groups using $L^*_{Align}$ as well as in the baseline $L_{TESSL}$. The role of temporal signal in pre-training is suggested by the decline in $T_0$ performance when excluding MTAN, in contrast to the full time-series performance (Fig. \ref{fig:results}.A).

Linear classification results demonstrate that the learned representation carry relevant information for pCR prediction, outperforming previous methods \cite{zhang2024m2fusion, jing2024prediction} (see Sec.\,\ref{sec:relatedwork}). Pre-training of representations using longitudinal data, also improves prediction using only single time point ($T_0$) and early response ($T_0+T_1$) predictions, surpassing reported results.

Although ISPY-2 is a multi-center dataset, further validation on independent datasets would enhance the generalizability of our findings. Additionally, 3D CNNs would be ideal for volumetric information, but memory constraints and the size of the dataset limited us to 2D MIPs.

\section{Conclusion}
Predicting pCR in breast cancer patients is challenging due to the heterogeneity of individual response behavior. This study demonstrates that representing temporal dynamics can improve prediction accuracy. It identifies response-specific patterns in imaging data by balancing reconstruction, temporal continuity, and alignment of responder time-series. Evaluated with a frozen encoder and linear classifier, our method outperformed both the $L_{TESSL}$ loss for time-series~\cite{thrasher2024te} and prior results reported on the ISPY-2 dataset \cite{jing2024prediction}, highlighting the effectiveness of the pre-training approach. 
Predicting pCR using the full time-series has the potential to inform forgoing surgery. Results show that prediction based on early response is feasible, offering a perspective for early therapeutic adjustment.

\begin{credits}
\subsubsection{\ackname} This work was funded by the Vienna Science and Technology Fund (WWTF, PREDICTOME [10.47379/LS20065]) the European Union’s Horizon Europe research and innovation programme under grant agreement No.101100633— EUCAIM, Austrian Federal Ministry of Labour and Economy, the National Foundation for Research, Technology and Development and the Christian Doppler Research Association, and Siemens Healthineers. 
\subsubsection{\discintname}The authors have no competing interests to declare that are relevant to the content of this article.
\end{credits}

%
%
%
\bibliographystyle{splncs04}

\end{document}